\documentclass[sigconf, anonymous=False]{acmart}
\setcopyright{acmcopyright}
\copyrightyear{2024}
\acmYear{2024}
\acmConference[]{Pre-print}{Under Review}{V 1.0}

\usepackage[normalem]{ulem}
\useunder{\uline}{\ul}{}
\usepackage{algorithm}
\usepackage{algorithmic}
\usepackage{multirow}
\usepackage{amsmath} 
\usepackage{enumitem}

\usepackage{caption} 

\usepackage{etoolbox} 

\begin{document}
\newcommand{\qa}[1]{{\color{black}#1}}

\title{Mitigating Entity-Level Hallucination in Large Language Models}

\author{Weihang Su}
\email{swh22@mails.tsinghua.edu.cn}
\affiliation{
Department of Computer Science and Technology, Tsinghua University,
Beijing 100084 \country{China}
}

\author{Yichen Tang}
\authornote{Contributed equally}
\affiliation{
Department of Computer Science and Technology, Tsinghua University,
Beijing 100084 \country{China}
}

\author{Qingyao Ai}
\authornote{Corresponding author}
\email{aiqy@tsinghua.edu.cn}
\affiliation{
Department of Computer Science and Technology, Tsinghua University,
Beijing 100084 \country{China}
}

\author{Changyue Wang}
\affiliation{
Department of Computer Science and Technology, Tsinghua University,
Beijing 100084 \country{China}
}

\author{Zhijing Wu}
\affiliation{
School of Computer Science and
Technology, Beijing Institute of Technology,
Beijing 100081 \country{China}
}

\author{Yiqun Liu}
\affiliation{
Department of Computer Science and Technology, Tsinghua University,
Beijing 100084 \country{China}
}

\begin{abstract}



The emergence of Large Language Models (LLMs) has revolutionized how users access information, shifting from traditional search engines to direct question-and-answer interactions with LLMs. 
However, the widespread adoption of LLMs has revealed a significant challenge known as hallucination, wherein LLMs generate coherent yet factually inaccurate responses.
This hallucination phenomenon has led to users' distrust in information retrieval systems based on LLMs.
To tackle this challenge, this paper proposes Dynamic Retrieval Augmentation based on hallucination Detection (DRAD) as a novel method to detect and mitigate hallucinations in LLMs. 
DRAD improves upon traditional retrieval augmentation by dynamically adapting the retrieval process based on real-time hallucination detection.
It features two main components: Real-time Hallucination Detection (RHD) for identifying potential hallucinations without external models, and Self-correction based on External Knowledge (SEK) for correcting these errors using external knowledge. 
Experiment results show that DRAD demonstrates superior performance in both detecting and mitigating hallucinations in LLMs.
All of our code and data are open-sourced at ~https://github.com/oneal2000/EntityHallucination.
\end{abstract}


\maketitle

\section{Introduction}

In recent years, large language models (LLMs) have achieved remarkable success in a variety of natural language processing (NLP) tasks and have already become an indispensable component of many AI applications~\cite{brown2020language,chowdhery2022palm,touvron2023llama,scao2022bloom,zhang2022opt}.
Due to the outstanding capabilities and wide applications of LLMs, they have revolutionized how users access information, shifting from traditional search engines to direct question-and-answer interactions with LLMs.
However, despite their impressive performance, it is well recognized that existing LLMs may generate text that, while appearing coherent and plausible on the surface, is fundamentally inaccurate or lacks grounding in reality.
This phenomenon is commonly referred to as LLM hallucination~\cite{maynez2020faithfulness,zhou2020detecting,liu2021token,ji2023survey,su2024unsupervised}.

To address hallucination, the most popular method adopted in existing studies is Retrieval-Augmented Generation (RAG). In this method, relevant knowledge is retrieved from an external corpus and utilized as input for LLMs, which has been proven to be effective in many NLP tasks~\cite{khandelwal2019generalization,borgeaud2022improving,lewis2020retrieval,su2024stard,jiang2022retrieval,su2024dragin}.  
Traditional RAG methods often adopt single-round retrieval that uses the original input of LLM as the query to retrieve relevant external information. 
Although this method is effective for simple tasks, it frequently fails in complex tasks like long-form generation and multi-hop question-answering. 
This is primarily because the LLM needs continuously changing information in these complex text generation tasks~\cite{jiang2023active}. 
In contrast, multi-round RAG~\cite{trivedi2022interleaving,borgeaud2022improving,ram2023context,jiang2023active} performs multiple retrievals during the generation process of LLMs. 
Depending on the timing of retrieval augmentation, various methods have been proposed in this area.
For example, RETRO~\cite{borgeaud2022improving} and IC-RALM~\cite{ram2023context} trigger the retrieval module based on a pre-defined number of generated tokens.
FLARE~\cite{jiang2023active} triggers the retrieval module whenever a token's predictive probability is below a certain threshold.
DRAGIN~\cite{su2024dragin} defines an empirical method based on uncertainty and self-attention to determine when to trigger retrieval.

Despite these advancements, a significant oversight remains: none of the existing studies explicitly verify whether the retrieval is triggered at an optimal timing. 
Typically, the evaluation methods of existing dynamic RAG approaches only focus on the performance in downstream QA tasks, without directly evaluating whether the chosen moment to invoke retrieval is appropriate.
In this paper, we argue that retrieval augmentation without identifying when and where hallucination happens in LLM is suboptimal in terms of effectiveness and efficiency.
On the one hand, the retriever cannot be perfect, therefore unnecessary retrieval augmentation may introduce irrelevant or noisy data to LLMs.
On the other hand, calling the retrieval module during the LLM generation process will increase the inference time and computational cost.
Such cost is unworthy if retrieval augmentation is conducted in places where LLM hallucination doesn't exist.

\begin{figure*}[t]
\centering
    \includegraphics[width=0.85\textwidth]{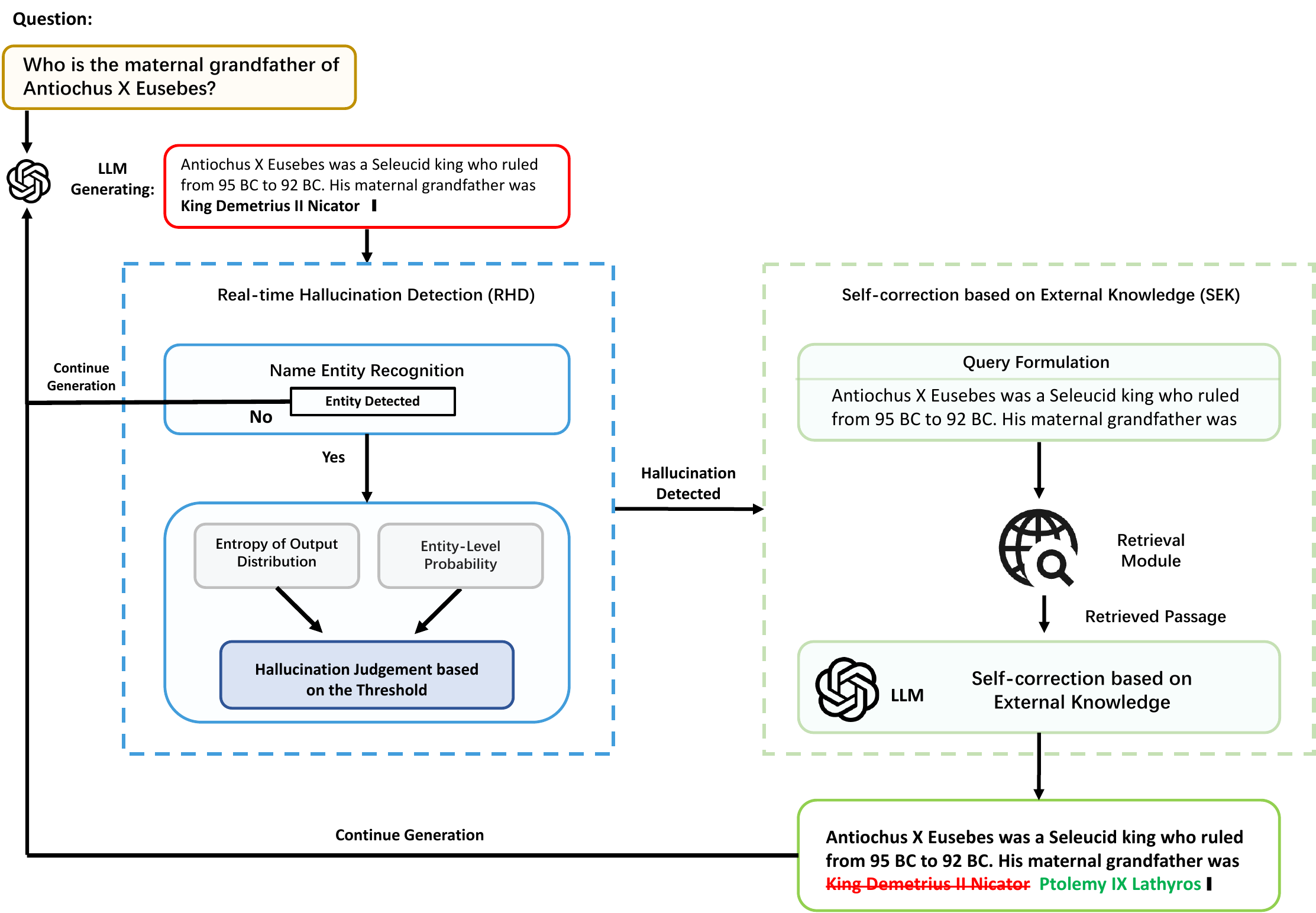}
    \caption{An illustration of our proposed DRAD Framework, which comprises two main components: RHD and SEK Module, highlighted in the diagram with blue and green frames, respectively.}
    \label{pic:big}
\end{figure*}

To address these concerns, we evaluate existing dynamic RAG approaches on the timing of retrieval to find out if it coincides with the occurrence of hallucination. 
We then propose a dynamic retrieval-augmented generation strategy based on hallucination detection. 
To be specific, we propose DRAD, Dynamic Retrieval augmentation based on hallucination Detection (\textbf{DRAD}, illustrated in Figure~\ref{pic:big}) that synchronizes retrieval augmentation with real-time hallucination detection of LLMs during the text generation process.
DRAD consists of two components: Real-time Hallucination Detection ({RHD}) and Self-correction based on External Knowledge ({SEK}). 
RHD is a real-time hallucination detection method that doesn't rely on any external models or external knowledge. 
It detects hallucinations by analyzing the uncertainty of the output entities of an LLM, particularly those with low probability and high entropy, that are likely to be potential hallucinations.
When RHD determines that the model is likely to generate unreliable text, SEK invokes the retrieval module to retrieve relevant external knowledge and helps the LLM to make corrections to its outputs so that potential hallucinations can be prevented.

We conducted experiments on existing benchmarks~\cite{manakul2023selfcheckgpt,ho2020constructing,stelmakh2022asqa,geva2021did,hayashi2021wikiasp} to evaluate the effectiveness of our framework. 
Experimental results show that RHD can achieve state-of-the-art (SOTA) performance in hallucination detection, and DRAD significantly outperforms existing single-round and multiple-round retrieval augmentation methods.

To summarize, the contributions of this paper are as follows:

\begin{itemize}[leftmargin=0.1\columnwidth]
\item We introduce a new retrieval-augmented framework, i.e., DRAD\footnote{Our code and data are open-sourced on this anonymous Github link:~{\url{https://github.com/oneal2000/EntityHallucination}}.}, which detects hallucinations during LLM's inference process and triggers RAG only when hallucinations are detected. 

\item 
We propose a real-time hallucination detection method, i.e., RHD, that achieves SOTA performance on existing benchmarks.

\item 
We evaluate DRAD on multiple complex QA benchmark datasets, and the experimental results demonstrate that our proposed DRAD framework significantly reduces hallucinations in large models across three diverse text generation benchmarks, outperforming previous methods.

\end{itemize}

\section{Related Works}

\subsection{Hallucination Detection}

Given the significance of hallucination detection and mitigation, considerable research has focused on developing efficient and effective methods for hallucination detection. 
Some methods require the model to generate multiple outputs for the same input. 
For instance, SelfCheckGPT~\cite{manakul2023selfcheckgpt} introduces three methods: SelfCheckGPT\_BERTScore, SelfCheckGPT\_QA, and SelfCheckGPT\_n-gram. These methods allow an LLM to generate multiple outputs based on the same input. Subsequently, the likelihood of hallucinations occurring in the LLM is measured based on the consistency among these outputs.
On the other hand, some methods require the introduction of new models. For example, MIND ~\cite{su2024unsupervised} is an unsupervised hallucination detection method based on the internal states of LLMs. The MIND framework consists of two steps: automatically generating training data and training a hallucination detector based on the hidden states of the selected LLM. 
The input to the hallucination detector is the hidden states of the LLM, and the output is a 0-1 label, representing whether the LLM is experiencing hallucinations.
~\citeauthor{liu2021token} ~\cite{liu2021token} specifically fine-tuned several models such as BERT~\cite{devlin2018bert}, RoBERTa~\cite{liu2019roberta}, and XLNet~\cite{yang2019xlnet} to detect hallucinations. 
The input to the Hallucination Detector is the output text of an LLM, and the output is also a 0-1 label indicating whether the LLM is experiencing hallucinations.
Our proposed hallucination detection method does not require generating multiple responses for the same input or introducing external information or models. 
Consequently, our approach excels in efficiency compared to existing works. 

\subsection{Retrieval-augmented Generation}
In recent studies, Retrieval-Augmented Generation (RAG) has been widely used to improve the performance of LLMs.
One of the most direct methods is single-round RAG~\cite{khandelwal2019generalization,borgeaud2022improving,lewis2020retrieval,guu2020retrieval,izacard2020leveraging,jiang2022retrieval,shi2023replug}, which utilize the initial input of the LLM to retrieve external knowledge. 
The retrieved external knowledge is subsequently integrated into the model's input.
Previous studies, like REPLUG~\cite{shi2023replug} and UniWeb~\cite{li2023web}, have explored using Language Model-based feedback and adaptive search engines for retrieval augmentation in single-round retrieval, enhancing predictions and selectively incorporating external knowledge.

Single-round RAG can be quite effective for straightforward tasks or situations where the user's information needs are well-defined.
However, for the tasks that involve generating extensive text such as long-form generation and multi-hop QA, searching for external knowledge based on the initial input cannot sufficiently address the information needs for LLM ~\cite{jiang2023active}. 
As a result, researchers have begun to explore multi-round retrieval augmentation. 
For example, RETRO~\cite{borgeaud2022improving} and IC-RALM~\cite{ram2023context} trigger retrieval every 4 to 32 tokens, while IRCot~\cite{trivedi2022interleaving} triggers retrieval whenever a new sentence is generated. 
Looking at it from another perspective, FLARE~\cite{jiang2023active} triggers retrieval when any token in the generated text has a probability lower than a certain threshold. 
However, indiscriminately considering the probability of every token is not the optimal solution for multi-round retrieval, as many tokens are function words (such as `am', `to', `in', etc) lacking semantic meaning.
To address the limitations of FLARE, DRAGIN~\cite{su2024dragin} optimize the timing (when to retrieve) and the query formulation method (what to retrieve) of the dynamic RAG framework, achieving state-of-the-art performance on downstream QA tasks.

\section{Methodology}
This section will introduce the DRAD framework in detail. DRAD consists of two components: Real-time Hallucination Detection ({RHD}) and Self-correction based on External Knowledge ({SEK}). We will introduce RHD in section ~\ref{sec:RED} and SEK in section~\ref{sec:SEK}.

\subsection{Real-time Hallucination Detection}\label{sec:RED}
\subsubsection{Empirical Investigation}
\begin{figure}[t]
\centering
    \includegraphics[width=0.8\columnwidth]{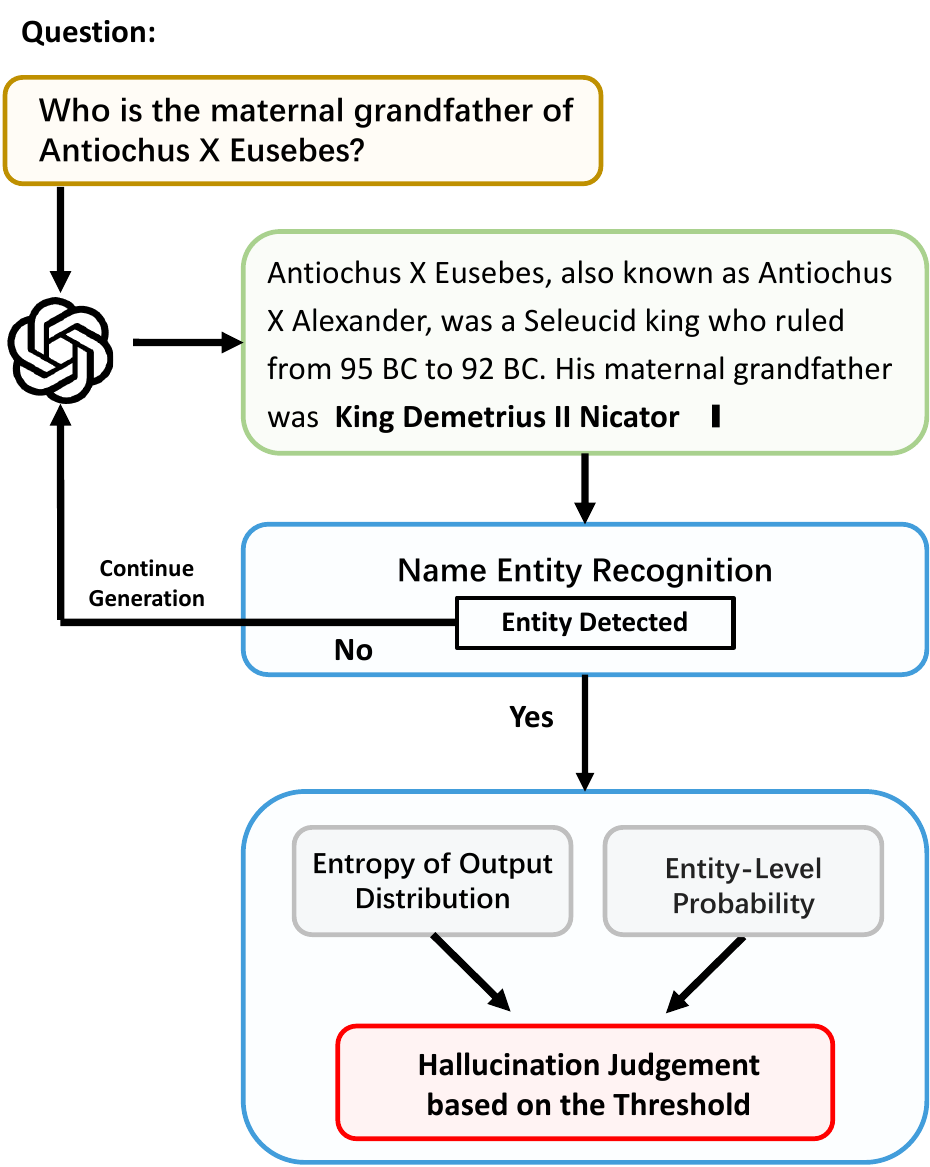}
    \caption{An illustration of our proposed Real-time Hallucination Detection (RHD) framework.}
    \label{pic:red}
\end{figure}

Our approach referred to as Real-time Hallucination Detection (RHD, illustrated in Figure~\ref{pic:red}), is constructed based on the assumption that when an LLM is forced to provide answers or generate text beyond its existing knowledge, the uncertainty in its output significantly escalates~\cite{manakul2023selfcheckgpt}. For example, consider the input, "Bill Clinton was born and raised in\textbf{ \_\_\_\_}".
Since the LLM has seen extensive information about Bill Clinton during the pre-training phase, it can confidently assign a high probability to the token ``Arkansas" to complete the sentence. 
Conversely, other locations such as ``Alabama" or ``Paris" would be considered low probability. 
However, in scenarios where the LLM is required to generate text about unfamiliar topics, such as "Alice's childhood neighbor now lives in\textbf{ \_\_\_\_}," the model faces a challenge. 
It lacks specific knowledge to confidently choose a word to fill the gap, leading to a flat probability distribution across all tokens related to place names in its vocabulary.
This uncertainty in choosing the right token can cause the model to select a random entity during text generation and cause hallucinations.

This insight leads us to establish a connection between uncertainty metrics and hallucination detection in text generation. 
When a model's internal parameters cannot reliably choose the correct response from multiple options, its outputs tend to include tokens with low probabilities and high entropy. 
Therefore, we propose a method for real-time detection of hallucinations in LLM's outputs, utilizing an entity confidence metric derived from both the predictive probability and entropy of entities.

\subsubsection{Entity Probability}\label{sec:prob}

The first method to evaluate the hallucination of an LLM involves examining its generation probability for each entity in its output. 
This process starts with the recognition of entities in the LLM output. 
Specifically, let \( O \) represent the output sequence generated by an LLM. We apply real-time entity recognition on \( O \) to identify entities. Assume an entity \( E \) is detected in \( O \), consisting of \( n \) tokens. Next, for each token \( t_i \) in \( E \), where \( 1 \leq i \leq n \), let \( p_i \) denote the generation probability of \( t_i \) as given by the output layer of the LLM.
Finally, the overall probability of the entity \( E \) is computed as an aggregation of the probabilities of its constituent tokens. This can be expressed as:

\begin{equation}
    P(E) = f(p_1, p_2, ..., p_n) ,
\end{equation}

\noindent where \( f \) is an aggregation function (e.g., product, sum, average, min, max) applied to the probabilities \( p_1, p_2, ..., p_n \) of the tokens in \( E \).

\subsubsection{Entity Entropy}\label{sec:entropy}

In addition to probability, \emph{entropy} is widely used to evaluate the uncertainty inherent in the model's output distribution.  Given an entity \( E \) comprised of \( n \) tokens, we assess the entropy for each token in the LLM output distribution.
For each token \( t_i \) in the entity \( E \), the entropy \( \mathcal{H}_i \) in the LLM output distribution is defined as:
\begin{equation}
    \mathcal{H}_i = -\sum_{\tilde{w} \in \mathcal{W}} p_i(\tilde{w}) \log p_i(\tilde{w}) ,
\end{equation}

\noindent where \( p_i(\tilde{w}) \) represents the likelihood of generating word \( \tilde{w} \), and \( \mathcal{W} \) is the vocabulary of the LLM.
To compute the overall entropy of the entity \( E \), we aggregate the entropies of its constituent tokens. This can be done using a pooling function \( f \) that takes the entropies of individual tokens and combines them to yield the entropy of the entity:

\begin{equation}
    \mathcal{H}(E) = f(\mathcal{H}_1, \mathcal{H}_2, ..., \mathcal{H}_n) .
\end{equation}

This approach, using entropy as a measure, offers an additional dimension to evaluate the uncertainty in an LLM's output, particularly in its ability to generate specific entities.

\subsubsection{Threshold for Hallucination Detection}

To explicitly determine whether an entity generation should be considered as a hallucination or not, we define two thresholds, \( \theta_1 \) and $\theta_2$.

\begin{equation}
\text{Entity Hallucination} =
\begin{cases}
\text{yes} & \text{if } P(E) < \theta_1 \text{ or } \mathcal{H}(E) > \theta_2 \\
\text{no} & \text{otherwise}
\end{cases}
\end{equation}

Note that \( \theta \) determines the frequency of calling the retrieval module. The specific value of \( \theta \) can be adjusted based on the demands of the actual application, thereby offering flexibility in determining how often the retrieval module is consulted. We conducted a detailed experiment to explore the impact of the threshold on efficiency and effectiveness. The experimental results can be found in section ~\ref{subsec:text-generation}.

\subsection{Self-correction based on External Knowledge}\label{sec:SEK}
The objective of the DRAD framework is to mitigate hallucinations in LLM using a Detect-Retrieve-Revise paradigm. 
The RHD module performs detection, which involves real-time identification of hallucinations. 
Following hallucination detection, DRAD retrieves relevant external knowledge based on the context surrounding the hallucination's occurrence and then revises the hallucination. 
This Retrieve-Revise process is implemented through the SEK method (Self-correction based on External Knowledge). 
The SEK method comprises three main components: query formulation, relevant knowledge retrieval, and self-correction. This section provides a detailed description of the SEK method.

\subsubsection{Formulating Search Queries}\label{subsec:searchQueries}
Given the output of an LLM \( O \) comprising tokens \( t_1, t_2, \ldots, t_n \) and an entity \( E \) identified as a hallucination by the RHD module, where the token span of \( E \) is from \( t_k \) to \( t_{k+i} \), the SEK query \( Q \) is formulated as:
\begin{equation}
Q = f(O, t_k, i) = \text{concat}(t_{k-m}, \ldots, t_{k-1}, t_{k+i+1}, \ldots, t_{k+m}) .
\end{equation}
In this equation, \( Q \) represents the formulated query, \( O \) denotes the output of the LLM consisting of a series of tokens, \( t_k \) to \( t_{k+i} \) define the hallucination entity span within the context, and \( \text{concat} \) is the concatenation function that assembles tokens surrounding the entity span to construct the query. The function \( f \) represents the process of query generation based on the given context and entity span. The parameter \( m \) is a flexible hyperparameter that determines the range of tokens included in the concatenation process.

\subsubsection{External Knowledge Retrieval}\label{subsec:knowledgeIntegration}

Given an external knowledge corpus \(\mathcal{D}\), composed of a set of documents \(\{d_1, d_2, \ldots, d_n\}\), and a retrieval function \(\mathcal{R}\), the process of external knowledge retrieval for a given query \(q\) is mathematically formalized as follows:

\begin{equation}
\mathcal{S} = \{\mathcal{R}(q, d_i) | d_i \in \mathcal{D}\}.
\end{equation}
In this equation, \(\mathcal{S}\) represents a score list where each score is computed by the retrieval function \(\mathcal{R}\) for the query-document pair \((q, d_i)\), assessing their relevance.

\begin{equation}
\mathcal{S}_{\text{sorted}} = \text{sort}_{desc}(\mathcal{S}),
\end{equation}
where \(\mathcal{S}_{\text{sorted}}\) is the list of scores \(\mathcal{S}\) sorted in descending order based on relevance. Finally, the top \(k\) documents from \(\mathcal{S}_{\text{sorted}}\) are selected as the set of retrieved relevant documents:

\begin{equation}
\label{eq:rel}
\mathcal{D}_{\text{retrieved}} = \{d_i | (q, d_i, s_i) \in \mathcal{S}_{\text{sorted}} \wedge i \leq k\}.
\end{equation}
In this equation, \(\mathcal{D}_{\text{retrieved}}\) denotes the subset of documents from \(\mathcal{D}\) that are most relevant to the query \(q\), based on the top \(k\) scores in \(\mathcal{S}_{\text{sorted}}\).

Note that in the field of information retrieval, various methods have been explored for the retrieval function $R$. These include vocabulary-based approaches such as  such as TF-IDF~\cite{ramos2003using}, BM25~\cite{robertson2009probabilistic}, and Query Likelihood~\cite{zhai2008statistical}, as well as neural network-based methods~\cite{ma2021prop,su2023wikiformer,fang2024scaling,su2023caseformer,su2023thuir2,ma2023caseencoder,ye2024relevance,li2023thuir,chen2023thuir,li2023thuir2,chen2022web,li2023towards}. 


\subsubsection{Self-Correction}\label{subsec:selfCorrection}

The RHD method detects the specific position in an LLM's output where hallucination occurs. It is important to note that the text preceding this hallucination position is typically accurate and does not require modification. Thus, to revise the hallucination tokens, the first step of Self-correction is truncating the output of an LLM at the position of hallucination.
This truncation process is mathematically represented as:
\begin{equation}
O' = \text{truncate}(O, t_{k}),
\end{equation}
\noindent where \( O' \) denotes the truncated output, \( O \) is the original output of the LLM, and \( t_{k} \) denotes the first token of the hallucinated entity. 
Following this, the retrieved external knowledge set \( \mathcal{D}_{\text{retrieved}} \) (defined in Eq~\ref{eq:rel}) is input into the LLM using a specific prompt template\footnote{The prompt template is detailed in Section ~\ref{sec:setting}}.
After the external knowledge is concatenated, the LLM regenerates the content at the position of the hallucination, utilizing the external knowledge.
The regenerated content represents the LLM's self-correction based on external knowledge, aiming to address the detected hallucinated entities in its output.

\subsection{Discussion}
In this subsection, we discuss the limitations of the RHD method. 
While the RHD method is effective at detecting hallucinations caused by gaps in the knowledge of LLMs, it is less successful at identifying hallucinations that are due to incorrect information acquired during the model’s pre-training phase.
Hallucinations in LLMs primarily arise from two sources: first, the absence of relevant knowledge, which leads the LLM to generate the most probable but potentially incorrect token; second, the learning of incorrect knowledge during pre-training. 
The RHD method utilizes uncertainty for hallucination detection, making it proficient at recognizing the first type of hallucination, where LLMs often show uncertainty. However, it struggles with the second type, where LLMs might deliver incorrect responses with high confidence.

\section{Experimental Setup}

\subsection{Datasets}

\subsubsection{Hallucination Detection}
We evaluate our proposed real-time hallucination detection method on {WikiBio GPT-3 dataset}~\cite{manakul2023selfcheckgpt}. This dataset comprises text generated by GPT-3 based on various topics, along with manually annotated labels indicating whether these passages contain hallucinations from a factual perspective.

\subsubsection{Retrieval-augmented Text Generation}
We evaluate the text generation performance of DRAD on the following datasets. 
\begin{itemize}[leftmargin=0.1\columnwidth]

    \item \textbf{2WikiMultihopQA}~\cite{ho2020constructing}. We use the 2WikiMultihopQA dataset to assess the DRAD model's ability to answer complex questions. 2WikiMultihopQA comprises complex questions that require two hops of information from Wikipedia articles. Answering these questions typically involves composing, comparing, or inferring multiple pieces of information to provide an accurate response.

    \item \textbf{StrategyQA} ~\cite{geva2021did}. We utilize the StrategyQA dataset to assess the commonsense reasoning capabilities of DRAD. StrategyQA comprises a diverse range of crowdsourced yes/no questions.

    \item \textbf{NQ} ~\cite{kwiatkowski2019natural}. We use the NQ dataset to assess the DRAD model's capability to answer factual questions. NQ is a large-scale question-answering dataset derived from real queries of Google Search.

\end{itemize}

\subsection{Settings for each Dataset}
\label{sec:setting}

Our settings on various benchmarks are as follows:

\begin{itemize}[leftmargin=0.1\columnwidth]

    \item \textbf{2WikiMultihopQA}~\cite{ho2020constructing}. We follow the prompt template of ~\citeauthor{wang2022self} ~\cite{wang2022self} that instructs the selected LLM to generate the chain-of-thought reasoning process. Following the prompt template of ~\cite{trivedi2022interleaving} and ~\cite{jiang2023active}, we add 8 example query-answer pairs to the input as prompts. 
    

    For the evaluation metrics, we use pattern-matching techniques to extract the final answer from the output of the LLM. This extracted answer is subsequently compared to the reference answer. 
    We employ various methods for this comparison, including the exact match (EM) metric at the answer level and token-level assessments of F1 score, precision, and recall.

    \item \textbf{StrategyQA} ~\cite{geva2021did}. We follow the setting of ~\citeauthor{wei2022chain} ~\cite{wei2022chain} to instruct the LLM to generate the chain-of-thought reasoning process. Following the prompt template of ~\cite{wei2022chain} and ~\cite{jiang2023active}, we add 8 example query-answer pairs to the input as prompts. 
    
    Since this dataset only contains true/false questions, we directly reported the accuracy.


    \item \textbf{NQ} ~\cite{kwiatkowski2019natural}. Since the questions in NQ are relatively simple, we did not include example query-answer pairs as prompts in the input to LLM. 
    
    For the evaluation metrics, we use the same parameters as we did in our 2WikiMultihopQA experiment.

\end{itemize}

For all three datasets, we chose BM25 as our retrieval model based on findings from ~\cite{ram2023context}, which demonstrated its superior performance in RAG, even outperforming SOTA dense retrieval models. 
For the external corpus, we use Wikipedia passages. The Top 3 relevant passages retrieved by BM25 were added to the prompt.



\subsection{Baselines}

\subsubsection{Hallucination Detection} We chose the following hallucination detection methods as baselines:

    \begin{itemize}[leftmargin=0.1\columnwidth]

        \item \textbf{SCG-BERTScore}~\cite{manakul2023selfcheckgpt} uses the LLM to produce multiple samples for a single input. BERTScore calculates the average similarity between a sentence and its most similar counterpart in various samples. If a sentence is highly similar to many samples, it's probably factual. If not, it might be a fabricated detail or "hallucination". 
        
        \item \textbf{SCG-QA}~\cite{manakul2023selfcheckgpt} generates multiple samples from the LLM for a single input. Based on one of these samples, a query generation system creates multiple-choice questions. Then a QA system uses the other samples to answer the question. The consistency of these answers is used to measure the probability of hallucination.

        \item \textbf{SCG-Ngram}~\cite{manakul2023selfcheckgpt} involves training an n-gram model, on samples generated by the selected LLM. As the sample size escalates, the behavior of this new model progressively approximates that of the original LLM. Subsequently, the average log probabilities for a given response, R, are computed based on the n-gram model. This average of log probabilities is then employed as a metric to quantify the likelihood of hallucination.

        \item \textbf{SCG-Ensemble}~\cite{manakul2023selfcheckgpt} is a simple combination of the normalized scores of SCG-BERTScore, SCG-QA, and SCG-Ngram.
        


        
        \item \textbf{Predictive Probability}~\cite{jiang2023active} involves leveraging the probabilities of tokens generated by LLMs as a metric to measure hallucination. FLARE~\cite{jiang2023active} adopts this simple methodology to determine the appropriate timing for invoking retrieval augmentation.

        \item \textbf{Predictive Entropy} utilizes the entropy of the output distribution of tokens generated in the text by large models as a metric to gauge hallucination.

    \end{itemize}

\subsubsection{Text Generation}
\begin{table}[t]
\caption{A comparative overview of our selected Retrieval-Augmented Generation baselines.}
\label{tab:baselines}
\begin{tabular}{lcc}
\toprule
              & \textbf{Timing for Retrieval}     & \textbf{Query Formulation}                                                                  \\
              \midrule
\textbf{SRR}  & Before Generation             & Initial Input                                                                      \\
\midrule
\textbf{FLR}  & Per Sentence            & Last Generated Sentence                                                                     \\
\midrule
\textbf{TPR}  & \begin{tabular}[c]{@{}c@{}}Any Token Probability\\ Below Threshold\end{tabular} & \begin{tabular}[c]{@{}c@{}}Last Generated Sentence\\ Exclude Uncertain Tokens\end{tabular} \\
\midrule
\textbf{DRAD} & \begin{tabular}[c]{@{}c@{}}Entity-based\\ Hallucination Detection\end{tabular} & Context of Hallucination           \\
\toprule
\end{tabular}
\end{table}

We choose the following Text Generation baselines for comparison. To ensure a fair comparison, each sentence in every multi-round RAG method can only be revised once.

\begin{itemize}[leftmargin=0.1\columnwidth]

\item  \textbf{NOR} (\textbf{NO R}etrieval). Directly generating answers without retrieval augmentation.
\item  \textbf{SRR} (\textbf{S}ingle-\textbf{R}ound \textbf{R}etrieval). In this setting, we retrieve relevant external documents once based on the initial question. Most of the current retrieval-augmented LLMs adopt this setting.
\item  \textbf{FLR} (\textbf{F}ix \textbf{L}ength \textbf{R}etrieval)~~\cite{trivedi2022interleaving}. A multi-round retrieval augmentation method that triggers the retrieval module for every sentence. 
\item \textbf{TPR} (\textbf{T}oken \textbf{P}robability \textbf{R}etrieval)~\cite{jiang2023active}. A multi-round retrieval augmentation method that triggers retrieval each time it encounters an uncertain token.


\end{itemize}

For multi-round Retrieval-Augmented Generation, the two most critical aspects are the timing for retrieval and the method of query formation when triggering retrieval. To visually demonstrate the differences between our selected baselines, we present the timing for retrieval and the method of query formation for each baseline in Table \ref{tab:baselines}.

\subsection{Implementation Details}\label{sec:implement}
In this subsection, we provide a comprehensive overview of our implementation details for the major components of our study: configurations for LLMs, Named Entity Recognition (NER), and the External Knowledge Corpus.

\begin{itemize}
\item \textbf{LLM Configuration:} We follow settings of ~\citeauthor{jiang2023active} ~\cite{jiang2023active} and validate our retrieval augmentation approach based on the GPT-3.5 language model, specifically the "text-davinci-003" variant, by iteratively querying its API~\footnote{https://api.openai.com/v1/chat/completions}. Notably, this API allows access to the probabilities of the generated tokens.

\item \textbf{Name Entity Recognition (NER):} For the Named Entity Recognition (NER) component of RHD, we follow the methodologies in prior studies~\cite{liu2021token,tarcar2019healthcare}. Specifically, we utilized the Spacy library, a tool recognized for its efficacy in NER as evidenced by previous research~\cite{shelar2020named}.

\item \textbf{External Knowledge Corpus:}
We adopt Wikipedia as our external knowledge corpus. Each article is segmented into 100-token passages. 
\end{itemize}

\begin{table}[t]

\caption{Experimental results of our hallucination detection technique compared to other baseline methods on the SeflCheckGPT dataset. The best performances are highlighted in bold. The performance of each SeflCheckGPT method is directly sourced from their original paper.}
\label{tab:hallucination}
\centering

\begin{tabular}{clc}
\toprule
                              &                      & \textbf{AUC}   \\
\midrule
\multirow{4}{*}{\textbf{Multi Generation}} & SCG\_BERTScore       & 81.96          \\
                              & SCG\_QA              & 84.26          \\
                              & SCG\_n-gram          & 85.63          \\
                              & SCG\_ensemble        & 87.33          \\
\midrule

\multirow{5}{*}{\textbf{Single Generation}} & Avg Token Entropy                & 80.73 \\
                               
                               & Avg Token Prob                & 83.21 \\
                               & Max Token Entropy                & 85.75 \\
                               
                               & Min Token Prob                & 87.51 \\
                             
                               & \textbf{RHD}(ours) & \textbf{89.31} \\
\toprule

\end{tabular}
\end{table}

\begin{table}[t]
\caption{Comparative experimental results for various token pooling methods of RHD. Since low probability and high entropy indicate low confidence, we employ min pooling for probability and max pooling for entropy.}
\label{tab:pooling}
\centering
\begin{tabular}{ccc}
\toprule
\multicolumn{1}{l}{} & \multicolumn{2}{c}{\textbf{AUC}} \\
\midrule
\textbf{Pooling Method}       & \textbf{Probability}  & \textbf{Entropy}  \\

\toprule
\textbf{Max}                  & -        & \textbf{88.00}    \\
\textbf{Min}                  & 88.75        & -   \\
\textbf{First}                & 87.60        & 87.46   \\
\textbf{Average}              & \textbf{89.31}        & 87.91    \\

\toprule
\end{tabular}
\end{table}


\begin{table*}[t!]
\caption{Experimental results of DRAD and other baselines on 2WikiMultihopQA, NQ, and StrategyQA. The best results are in bold. \#Num indicates the times of retrieval module calls, smaller means more efficient.}
\label{tab:all}
\begin{tabular}{llccccccccc}
\toprule
                                                &                      & \multicolumn{5}{c}{\textbf{2WikiMultihopQA}}                                                      & \multicolumn{2}{c}{\textbf{NQ}}              & \multicolumn{2}{c}{\textbf{StrategyQA}}         \\
                                                \midrule
                                                &                      & \textbf{F1}     & \textbf{EM}   & \textbf{Prec.}  & \textbf{Recall} & \textbf{\#Num} & \textbf{F1}    & \textbf{\#Num} & \textbf{Accuracy} & \textbf{\#Num} \\
                                                \toprule
\textbf{Without Retrieval}                      & \textbf{NOR}         & 0.2939& 0.22          & 0.2879          & 0.3001          & 0                           & 0.283          & 0                           & 0.64              & 0                           \\
\midrule
\textbf{Single-round Retrieval}                 & \textbf{SRR}         & 0.3768& 0.26          & 0.3622          & 0.3927          & \textbf{1}                           & 0.293          & 1                           & 0.65              & 1                           \\
\midrule
\multirow{3}{*}{\textbf{Multi-round Retrieval}} & \textbf{FLR}       & 0.4788& 0.38          & 0.4693          & 0.4886          & 9.61                        & 0.291          & 3.41                        & 0.62              & 5.48                        \\
                                                & \textbf{TPR}       & 0.3866& 0.24          & 0.3681          & 0.4071          & 5.29                        & 0.252          & 1.67                        & 0.60              & 2.25                        \\
                                                & \textbf{DRAD} & \textbf{0.4856}& \textbf{0.39} & \textbf{0.4741} & \textbf{0.4976} & 1.40                        & \textbf{0.339} & \textbf{0.48}                        & \textbf{0.76}     & \textbf{0.53}    \\
                                                \toprule
\end{tabular}
\end{table*}

\section{Experimental Results}

\subsection{Hallucination Detection}

In this section, we present the experimental findings obtained from the WikiBio GPT-3 dataset. This dataset has been specifically crafted for a binary classification task, with the objective of measuring the likelihood of hallucinated text produced by the GPT3 (text-davinci-003) model. The AUC (Area Under the Curve) metric was adopted to evaluate the performance of each hallucination detection method. 

\subsubsection{Overall Results}

The experiment results of RHD and other baselines are shown in Table~\ref{tab:hallucination}. We have the following observations. 
(1) Firstly, methods based on token probability are shown effective in detecting hallucinations. Among these, the minimum token-level probability has superior performance among all token-generation probability techniques. However, the performance of token-level hallucination detection methods generally falls short when compared to RHD. \textbf{This also confirms our assumption that indiscriminately considering the probability of every token to detect hallucinations is unreasonable, as many tokens are meaningless words.}
(2) Furthermore, the SCG\_ensemble method, while effective among multi-round generation techniques, has drawbacks due to the need for LLMs to create multiple responses, making it impractical for real-time use. Despite its high computational complexity, its performance advantage over probability-based methods is not substantial. Thus, we recommend using the token probability methods when possible, due to their efficiency and effectiveness.
(3) Lastly, our proposed hallucination detection method RHD outperforms all other baselines. This method enables real-time detection of hallucinations and exhibits excellent performance, providing a solid foundation for our proposed Detect-Retrieve-Revise paradigm.

\subsubsection{Ablation Study on Pooling Methods}\label{subsec:pooling}

In the output generated by LLM, numerous entities are composed of multiple tokens. Thus, we explore diverse pooling strategies for these entities that encompass multiple tokens which are shown in Table~\ref{tab:pooling}. Through the experimental results, we have the following observations: Firstly, different pooling methods have minimal impact on our RHD method, suggesting that RHD is not sensitive to pooling strategies. Additionally, the average pooling of Probability achieves the highest AUC with a score of 89.31, indicating it's the most effective method for representing an entity. The following experiments in this section all follow this setting.

\subsection{Downstream Question-Answering Tasks}\label{subsec:text-generation}

\begin{figure}[t]
\centering
    \includegraphics[width=\columnwidth]{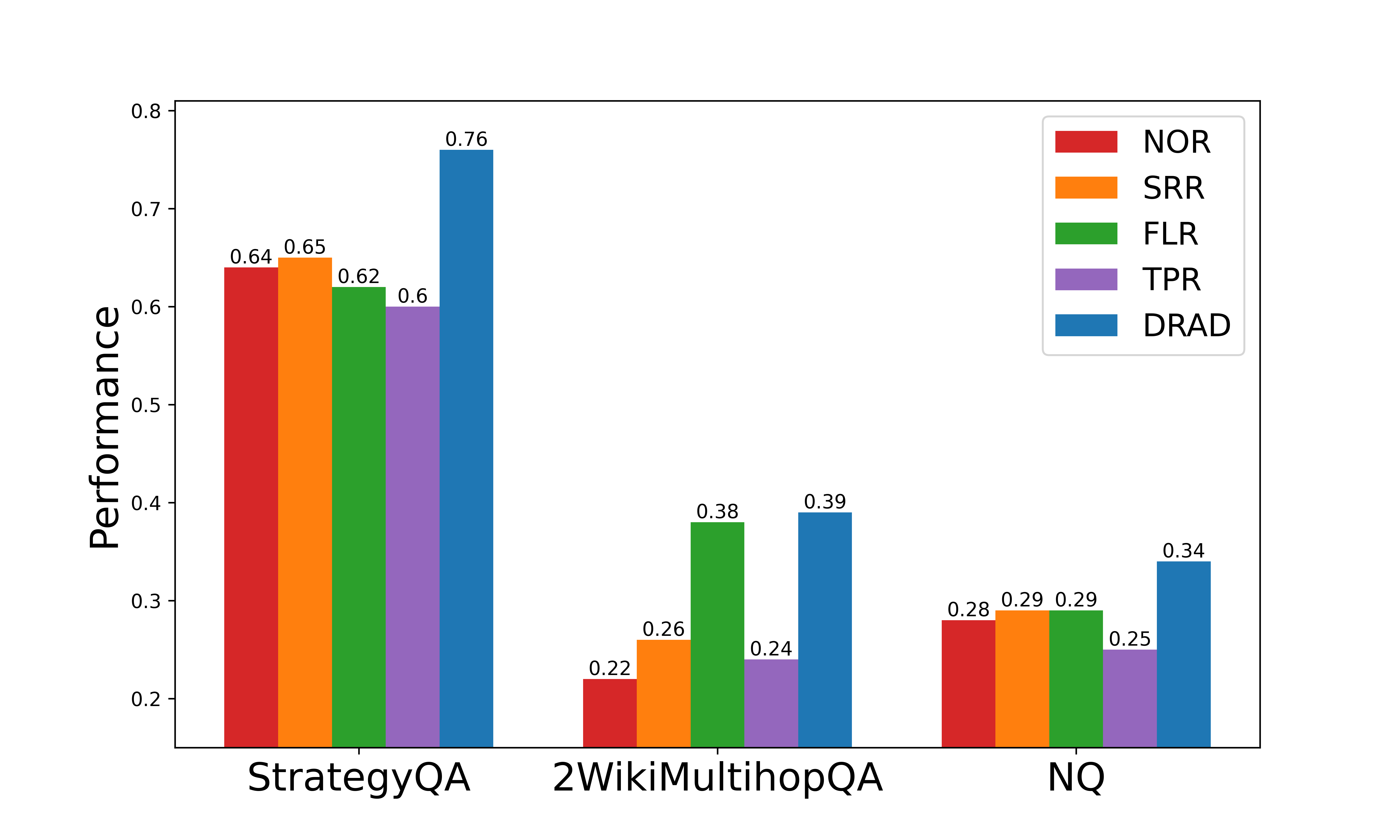}
    \caption{Visual comparison between DRAD and baselines across all datasets. For the evaluation metric, we choose Accuracy for StrategyQA, EM for 2WikiMultihopQA, and F1 for NQ.}
    \label{fig:all}
\end{figure}

\subsubsection{Overall Performance}

The overall experimental results of DRAD and other baselines\footnote{The implementation method of the TPR (FLARE) official Github repository is as follows: as long as the probability of all the tokens generated by LLM is not higher than the threshold, retrieval is continuously triggered to modify this sentence until the condition is met. To ensure a fair comparison, we stipulate that each sentence in every multi-round RAG method can only be revised once. That's why the performance reported in the original FLARE paper is higher than we reproduce.
} are presented in Table~\ref{tab:all} and Figure~\ref{fig:all}. 
The experimental results show that DRAD outperforms all baseline methods across various benchmarks. 
Specifically, in the context of StrategyQA and NQ datasets, DRAD demonstrates a remarkable improvement in performance relative to preceding retrieval-augmented models. 
For the 2WikiMultihopQA dataset, while the results are commensurate with those of FLR, it is noteworthy that the number of retrieval invocations for DRAD constitutes only 14.7\% of those necessitated by FLR. 
For the FLR method, though the retrieval module is invoked for every individual sentence, our experimental results suggest that such an approach does not invariably enhance the performance of Language Learning Models (LLMs). 
Particularly, when tested on the StrategyQA dataset, the performance of the FLR method was found to be subpar in comparison to the single-round retrieval methods. 
Additionally, despite utilizing the identical hyperparameters as delineated in the original paper on TPR, the performance of the TPR model was observed to be inferior to other baseline models. 
On another note, on datasets such as NQ and StrategyQA, the DRAD model necessitated even fewer invocations of the retrieval module than methods supplemented by single-round retrieval. 
This observation emphatically underscores the superior efficiency and effectiveness of our proposed methodology.

\subsubsection{Ablation Studies on Threshold}

This section presents the results of our ablation study on different thresholds. The comparison between different thresholds of RHD ($\theta_1$) on 2WikiMultihopQA and StrategyQA is shown in Table~\ref{tab:threshold}. 
The experimental results demonstrate that our method is not sensitive to hyperparameters. There is no significant performance difference in the range of 0.4 to 0.5.

It's important to note that the threshold directly determines the frequency of invoking the retrieval module. A higher threshold results in fewer calls to the retrieval. In practical applications, the threshold can be adjusted based on real-world requirements to balance efficiency and effectiveness.

\begin{table}[t]
\centering
\caption{Comparasion between different thresholds of RHD on 2WikiMultihopQA and StrategyQA. The best results are in bold.}
\label{tab:threshold}
\begin{tabular}{cccc}
\toprule
                   & \multicolumn{2}{c}{\textbf{2WikiMultihopQA}} & \textbf{StrategyQA} \\
                   \midrule
\textbf{Threshold} & \textbf{F1}            & \textbf{EM}         & \textbf{Accuracy}   \\
\toprule
\textbf{0.40}       & \textbf{0.4732}        & \textbf{0.39}                & \textbf{0.76}       \\
\textbf{0.42}       & 0.4708                 & \textbf{0.39}                & 0.75                \\
\textbf{0.44}       & 0.4682                 & \textbf{0.39}                & 0.75                \\
\textbf{0.46}       & 0.4715                 & 0.38                &\textbf{ 0.76}                \\
\textbf{0.48}       & 0.4689                 & 0.38                & 0.75                \\
\textbf{0.50}       & 0.4654                 & 0.38                & 0.75                \\

\toprule
\end{tabular}
\end{table}

\subsubsection{Efficiency of DRAD}
This section presents the experimental results regarding the efficiency of various multi-round retrieval-augmented LLMs as shown in Table~\ref{tab:eff}. We compared the number of retrieval calls made by SRR, FLR, TPR, and DRAD across different datasets. The results indicate that FLR, due to its requirement of invoking retrieval for every sentence, is the least efficient. For TPR's efficiency lies between that of DRAD and FLR. DRAD, however, demonstrates the fewest retrieval calls, making it the most efficient.

It is noteworthy that DRAD makes fewer than one retrieval call on the NQ and StrategyQA datasets, yet its performance surpasses that of Single-time retrieval. This underscores the superiority of the DRAD approach over other retrieval-augmented models in both efficiency and effectiveness.
\begin{table}[t]
\caption{Comparison between the efficiency of DRAD and other baselines across all datasets. The most efficient results are in bold. 2WMQA indicates the 2WikiMultihopQA dataset.}
\label{tab:eff}
\centering
\begin{tabular}{lcccc}
\toprule
                         & \multicolumn{4}{c}{\textbf{\#Num of retrieval}}        \\
                         \midrule
\textbf{Dataset}         & \textbf{FLR} & \textbf{TPR} & \textbf{DRAD}& \textbf{SRR} \\
\toprule
\textbf{NQ}              & 3.41           & 1.67           & \textbf{0.48}   & 1     \\
\textbf{2WMQA} & 9.61           & 5.29           & 1.40    & \textbf{1}   \\
\textbf{StrategyQA}      & 5.48           & 2.25           & \textbf{0.53}     & 1   \\
\toprule
\end{tabular}
\end{table}

\section{Conclusions and Future Works}

In this study, we introduce DRAD to mitigate hallucinations of LLMs by integrating real-time hallucination detection (RHD) and self-correction based on external knowledge (SEK). RHD monitors the LLM's output for potential hallucinations without external models, while SEK retrieves relevant information to adjust the LLM's output, preventing hallucinations. Experiments demonstrate DRAD's superiority in hallucination detection and its effectiveness compared to other retrieval augmentation methods.
We acknowledge the limitations of this paper, notably that the real-time hallucination detection method (RHD) depends on token probability data, which is unavailable from certain APIs. Future work aims to develop more real-time hallucination detection methods to overcome this constraint.


\bibliographystyle{ACM-Reference-Format}
\bibliography{sample-base}

\end{document}